%% file: TemporalConditionDSR.tex
\documentclass[pdflatex,sn-mathphys-num]{sn-jnl}%

\usepackage{graphicx}%
\usepackage{multirow}%
\usepackage{amsmath,amssymb,amsfonts}%
\usepackage{amsthm}%
\usepackage{mathrsfs}%
\usepackage[title]{appendix}%
\usepackage{xcolor}%
\usepackage{textcomp}%
\usepackage{manyfoot}%
\usepackage{booktabs}%
\usepackage{listings}%
 
\input{head.tex}

\begin{document}
 
\title[Article Title]{MR-ImagenTime: Multi-Resolution Time Series Generation through Dual Image Representations
}

\author[1]{\fnm{Xianyong} \sur{Xu}}
\author[1]{\fnm{Yuanjun} \sur{Zuo}}
\author[1]{\fnm{Zhizhong} \sur{Huang}}
\author[2]{\fnm{Yihan} \sur{Qin}}
\author[2]{\fnm{Haoxian} \sur{Xu}}
\author*[2]{\fnm{Leilei} \sur{Du}}
\author[2]{\fnm{Haotian} \sur{Wang}}

\affil[1]{
    \orgdiv{Department}, 
    \orgname{State Grid Hunan Electric Power Company Limited Research Institute \& Hunan Province Engineering Technology Research Center of Electric Power Multimodal Perception and Edge Intelligence}, 
    \orgaddress{\city{Changsha}, \state{Hunan}, \country{China}}
}
\affil[2]{
    \orgdiv{College of Computer Science}, 
    \orgname{Hunan University}, 
    \orgaddress{\city{Changsha}, \state{Hunan}, \country{China}}
}


\input{body/0.abstract.tex} 

\maketitle

    *Corresponding author. \\
    E-mail: leileidu@hnu.edu.cn \\
    Contributing authors: \\
    93615073@qq.com; hjzuoyuanjun@163.com; \\
    zhihong\_huang111@163.com; yihan\_qin@hnu.edu.cn; \\
    xuhaoxian13@163.com; wanghaotian@hnu.edu.cn

\input{body/1.introduction.tex}

\input{body/2.relatedWork.tex}

\input{body/3.problemDefinition.tex}
\input{body/4.proposedSolution1.tex}

\input{body/5.experimentalStudy.tex}

\input{body/6.conclusion.tex}
\input{body/7.ack.tex}

\bibliography{references/add} 
\input{body/8.Appendix.tex}

\end{document}

%% file: head.tex
\usepackage{booktabs} 
\usepackage{color} 
\usepackage{balance} 
\usepackage{url} 
\usepackage[all]{nowidow}
\usepackage{tabularx}
\usepackage{siunitx} 
\usepackage{tabularx,subfigure,multirow, graphicx}
\usepackage{epsfig}
\usepackage{epstopdf} 
\usepackage{enumitem}
\usepackage{booktabs}  
\usepackage{amsmath}
\usepackage{float}

\usepackage{amsthm,amsmath}  

\newcolumntype{L}[1]{>{\raggedright\arraybackslash}p{#1}}
\newcolumntype{C}[1]{>{\centering\arraybackslash}p{#1}}
\newcolumntype{R}[1]{>{\raggedleft\arraybackslash}p{#1}}

\usepackage[linesnumbered,ruled,vlined]{algorithm2e}
\SetKwRepeat{Do}{do}{while}
\SetCommentSty{mycommfont}

\long\def\comment#1{}

\newcommand{\nop}[1]{}

\theoremstyle{remark}

\theoremstyle{definition}

\newcommand{\solutionSimpleName}{MR-CDM}

%% file: body/0.abstract.tex
\abstract{
Time series forecasting is vital across many domains, yet existing models struggle with fixed-length inputs and inadequate multi-scale modeling. We propose MR-CDM, a framework combining hierarchical multi-resolution trend decomposition, an adaptive embedding mechanism for variable-length inputs, and a multi-scale conditional diffusion process. Evaluations on four real-world datasets demonstrate that MR-CDM significantly outperforms state-of-the-art baselines (e.g., CSDI, Informer), reducing MAE and RMSE by approximately 6–10 to a certain degree. 
}

\keywords{Time series forecasting, Multi-resolution decomposition, Diffusion model}

%% file: body/1.introduction.tex
\section{Introduction}

Time series data is fundamental to decision-making in domains such as finance, healthcare, and transportation. While deep learning models like RNNs, CNNs, and Transformers have advanced time series forecasting, recent work has explored diffusion models for their ability to capture complex temporal distributions. However, real-world series often exhibit multi-scale patterns (trends, seasonality, noise) and variable lengths, posing challenges for standard diffusion approaches. Although trend-decomposition methods improve multi-scale modeling, they typically assume fixed input lengths. To address variable-length inputs, Naiman et al.~\cite{DBLP:conf/nips/NaimanBPAFA24} propose mapping time series to 2D image-like representations. While this enhances input flexibility, it risks distorting temporal continuity and long-range dependencies by treating sequential data as spatial grids.

To address these limitations, we propose a diffusion-based forecasting framework that (1) supports variable-length time series without fixed input windows, and (2) incorporates multi-scale trend decomposition to model temporal patterns at different resolutions. The method retains the generative strengths of diffusion models while improving robustness across heterogeneous sequence lengths and temporal scales. Experiments on multiple real-world datasets show that it consistently outperforms both diffusion-based and conventional baselines, achieving more accurate and reliable forecasts.


The remainder of this paper is organized as follows: 
Section~\ref{sec:related} reviews related literature on time series forecasting and diffusion-based generative models. 
Section~\ref{sec:ps} introduces the problem statement.
Section~\ref{sec:solution} presents the proposed \solutionSimpleName{} framework in detail. 
Section~\ref{sec:experiments} outlines the experimental setup and reports empirical the results. 
Finally, Section~\ref{sec:conclusion} concludes the paper and discusses potential directions for future research.

%% file: body/2.relatedWork.tex
\section{Related Work}\label{sec:related}
\renewcommand{\thesubsection}{\Alph{subsection}}   
\renewcommand{\thesection}{\arabic{section}}     

Traditional time series forecasting relies on statistical modeling methods based on statistical assumptions and linear models. ARIMA~\cite{DBLP:conf/nips/NaimanBPAFA24}, SARIMA, and VAR~\cite{ZhangZQ23} offer interpretable and multivariate-applicable structures, while Holt–Winters~\cite{KOEHLER2001269}, ETS~\cite{RePEc2009}, and Gaussian Processes~\cite{2005Gaussian} capture trend, seasonality, and Bayesian uncertainty quantification. With the deep learning revolution, neural networks enable powerful nonlinear modeling and hierarchical feature learning: LSTM~\cite{HSJ1997} and GRU~\cite{2014Learning} capture long-term dependencies and support the probabilistic model DeepAR~\cite{2020DeepAR}, WaveNet~\cite{2016WaveNet} and TCNs~\cite{DBLP:conf/iclr/BaiKK19} improve efficiency via parallel computing, and Transformer~\cite{2017Attention}, Informer~\cite{2020Informer}, and Autoformer~\cite{2021Autoformer} have become mainstream in capturing global temporal dependencies. Recently, the diffusion paradigm has emerged as an effective generative approach for time series, represented by TimeGrad~\cite{2021Autoregressive}, CSDI~\cite{2021CSDI}, SSSD~\cite{alcaraz2022diffusion}, TSDiff~\cite{kollovieh2023predict}, and DiffSTG~\cite{wen2023DiffSTG}, which achieve autoregressive denoising, conditional generation, efficiency enhancement, variable-length sequence processing, and spatio-temporal dynamics modeling.

%% file: body/3.problemDefinition.tex
\section{Problem Statement}\label{sec:ps}

\textbf{Problem Statement.}
%
We address the problem of time series forecasting, where the goal is to predict future values of a time series based on its past observations. Given a set of observed time series data $x \in \mathbb{R}^{L \times K}$, where $L$ is the sequence length and $K$ denotes the number of features, the challenge is to accurately model the complex temporal dependencies within the data, which span multiple time scales. This includes capturing the varying trends at different temporal resolutions and handling the high-dimensional and dynamic nature of the time series data for reliable forecasting.

%% file: body/4.proposedSolution1.tex
\section{Method}\label{sec:solution}

\begin{figure*}[t]
    \centering           
    \includegraphics[width=1\textwidth]{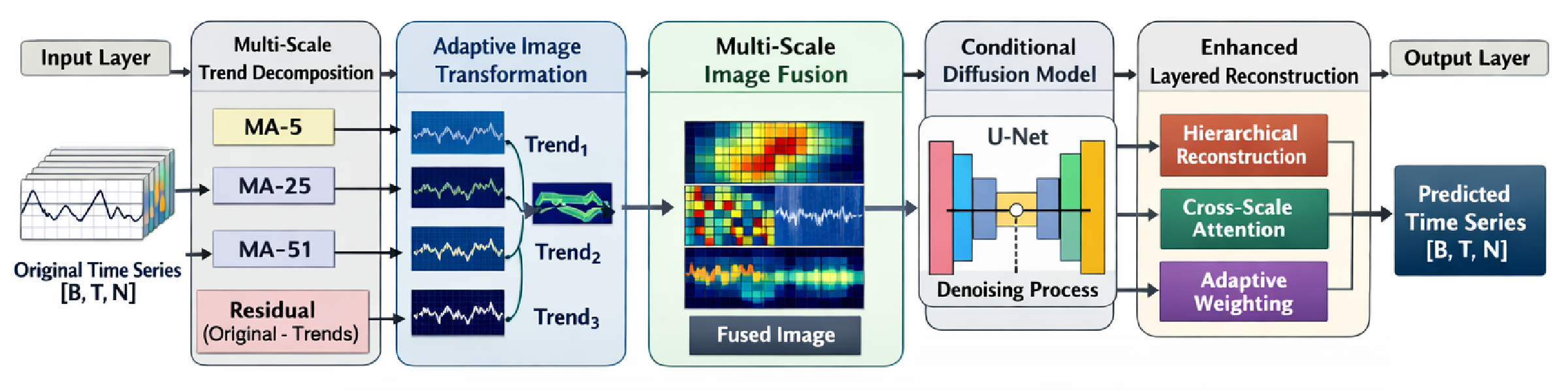}
    \caption{MR-CDM Model Architecture.}
    \label{fig:architecture}
\end{figure*}

The proposed MR-CDM framework follows a structured multi-stage pipeline for multi-scale time series modeling, as illustrated in \autoref{fig:architecture}. It first decomposes the input time series into short-term fluctuations, medium-term cycles, long-term trends, and high-frequency residuals using three-level multi-scale moving average decomposition with window sizes 5, 25, and 51~\cite{shen_multi-resolution_2024}. These components are then mapped into different domains: short- and medium-term signals are converted into 32×32 images via delay embedding to retain fine-grained temporal dependencies, while the long-term trend is transformed into the time-frequency domain using STFT to capture periodic amplitude and phase information~\cite{takens2006detecting,DBLP:conf/icassp/GriffinL83}. The resulting multi-branch representations are fused and fed into a conditional diffusion model~\cite{NEURIPS2020_4c5bcfec}, which performs generation conditioned on historical contexts to maintain temporal consistency. Finally, an Enhanced Hierarchical Reconstructor with cross-scale attention and adaptive weighting integrates multi-path information to recover high-fidelity predictions. This branch-wise design effectively decouples multi-scale temporal features, avoids low-frequency masking of high-frequency details, and leverages the strong generative capability of diffusion models for complex dynamic modeling.

%% file: body/5.experimentalStudy.tex
\section{Experiments}
\label{sec:experiments}

\subsection{Experimental Setup}



Detailed experimental settings are presented in Appendix~\ref{sec:Experimental Env} in our paper.


We adopt three primary evaluation metrics: Mean Squared Error (MSE), Mean Absolute Error (MAE), and Root Mean Squared Error (RMSE), which measure the squared error, absolute error, and scale-consistent error between predictions and ground truth, respectively.

\subsection{Datasets and Preprocessing}

\subsubsection{Dataset and Preprocessing}
We evaluate on the univariate LUFL series of the ETTh1 dataset (July 2016--July 2018), which contains 17,420 hourly time points. Following standard practice, missing values are imputed via linear interpolation and outliers are handled using the $3\sigma$ rule. The series is normalized using Z-score statistics from the training set and split chronologically into train (70\%), validation (10\%), and test (20\%) sets.Results on other domains are provided in Appendix~\ref{sec:Generalization} of our paper.


\subsubsection{Dataset Sythesis}

To demonstrate the generalization capability of our model, we generated a synthetic dataset, whose generation details are provided in the Appendix~\ref{sec:DS} in our paper.



\subsection{Baseline Models}

We employ three representative baselines across traditional statistical, deep learning, and advanced generative paradigms.For conventional time series modeling, we use ARIMA(2,1,2)~\cite{LiuY24} as a univariate baseline, which achieves 96-step prediction through linear trend extrapolation with Gaussian noise for uncertainty estimation.As a deep learning baseline, a two-layer stacked LSTM (128 hidden units, 0.2 dropout) is adopted, producing 96-step forecasts from the final hidden state and trained with AdamW for 200 epochs with cosine annealing and gradient clipping~\cite{2012-385}.For the state-of-the-art generative baseline, we select CSDI, a conditional score-based diffusion model that uses a masking strategy for conditional forecasting and 6 residual blocks integrating causal convolutions and 4-head self-attention to capture multi-scale temporal dependencies~\cite{TashiroSSE21}.

\subsection{Implements Details of Feature Processing}
Detailed implementation of feature processing is described in Appendix~\ref{sec:FP} in our paper.



\subsection{Main Results}

\subsubsection{Overall Performance Comparison}

Table~\ref{tab:main_results} reports the forecasting performance on the ETTh1 LUFL task on these models above.

\begin{table}[h]
\centering
\caption{Performance comparison on ETTh1.}
\label{tab:main_results}
\begin{tabular}{lcccc}
\hline
Model & MSE & MAE & RMSE\\
\hline
ARIMA & {42.2121} & {6.2893} & {6.496}\\
LSTM & {13.4981} & {3.6270} & {3.6739}\\
CSDI & {1.5538} & {0.8992} & {1.2465}\\
MR-CDM & {1.4842} & {0.9650} & {1.2183}\\
\hline
\end{tabular}
\end{table}



As shown in Table~\ref{tab:main_results}, we compare MR-CDM with two representative baselines and an advanced diffusion model on the ETTh1 feature prediction task. MR-CDM achieves superior performance, with a 96.5 percent MSE reduction compared to ARIMA and an 89.0 percent improvement over LSTM. Although CSDI outperforms ARIMA and LSTM, our model further surpasses CSDI. These results indicate that MR-CDM more effectively captures complex temporal patterns, highlighting the importance of explicitly modeling hierarchical trends.


The analysis of prediction result on ETTh1 using MR-CDM model is in Appendix~\ref{sec:aqc1} in our paper.

We also evaluate with advanced diffusion model CSDI. The results are in Appendix~\ref{sec:aqc} in our paper.

We also evaluate on a synthetic dataset. The results are in Appendix~\ref{sec:additional} in our paper.

A rigorous multiple-run validation protocol was employed to mitigate the impact of random variation and ensure fair comparison.Appendix~\ref{sec:multiple} in paper presents the comprehensive statistical summaries, confirming the reliability of our experimental setup and the consistent performance patterns across runs.

\begin{figure}[htbp]
	\centering           
	\includegraphics[width=1.0\linewidth]{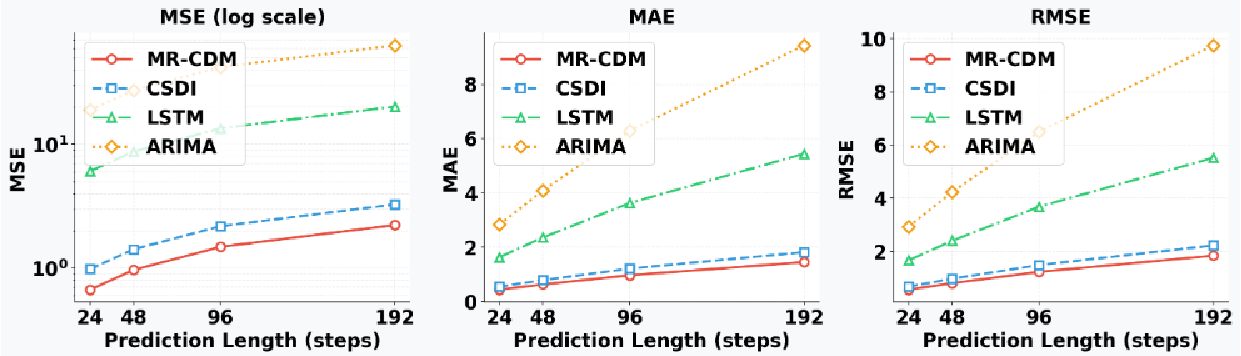}
	\caption{Multi-Step Prediction Performance Comparison.}
	\label{fig:multisteps}
\end{figure}

To evaluate the multi-step forecasting performance of MR-CDM on time series data, we conduct a comprehensive comparison experiment on the ETTh1 dataset. The historical input window is fixed at 96 time steps, while the prediction horizon is varied across four settings: 24, 48, 96, and 192 steps, enabling a systematic assessment of model accuracy and error degradation under increasing forecasting difficulty, as shown in \autoref{fig:multisteps}. Three baseline models are selected for comparison: ARIMA, LSTM, and CSDI. As illustrated in the figure, MR-CDM consistently achieves the lowest error across all prediction horizons, and its error growth rate remains significantly lower than that of the baselines as the prediction length increases. These results demonstrate that MR-CDM effectively suppresses error accumulation in long-range forecasting scenarios, confirming the superiority of the proposed approach for multi-step time series prediction tasks.

We conduct ablation studies to analyze the effectiveness of each component and input length sensitivity. Due to space limitations, please refer to Appendix~\ref{sec:ablation1}, Appendix~\ref{sec:ablation2} and Appendix~\ref{sec:ablation3} in our paper.

%% file: body/6.conclusion.tex
\section{Conclusion}
\label{sec:conclusion}

This thesis presents a comprehensive study on time series forecasting through the development of \textbf{MR-CDM}, a novel framework that addresses key limitations of existing methods. The research makes several significant contributions: (1) a delay embedding technique that converts variable-length time series into structured 2D image representations while preserving temporal dependencies, thereby enabling the use of spatial inductive biases from computer vision; (2) a hierarchical trend decomposition module that explicitly captures multi-scale temporal patterns—including short-term fluctuations, seasonal cycles, and long-term trends; and (3) a hierarchical conditional diffusion model that performs denoising generation under multi-scale guidance, reducing stochasticity through coarse-to-fine constraints.


    

%% file: body/7.ack.tex
\section*{Acknowledgment}
\label{sec:ack}
This research was funded by Science and Technology Project of State Grid Hunan Electric Power Company Limited, titled "Research on Key Technologies and Complete Equipment of Diffusion Super-Resolution Data Augmentation for Power Grid Model Identification and Situation Deduction", grant number 5216A5250009.

%% file: body/8.Appendix.tex
\appendix
\setcounter{secnumdepth}{2}
\setcounter{subsection}{0}

\renewcommand{\thesubsection}{\thesection.\arabic{subsection}}

\section{Ablation Studies}

\subsection{First Ablation Study}
\label{sec:ablation1}
To validate the effectiveness of key components in MR-CDM, we conduct ablation studies on the LUFL univariate forecasting task. We evaluate three variants:
(1) \textbf{Baseline-NoDecomposition}, which removes the multi-scale trend decomposition;
(2) \textbf{UnconditionalDiffusion}, which disables historical conditioning in the diffusion process; and
(3) \textbf{NoImageFusion}, which replaces our smart fusion module with simple feature concatenation.
These components are selected because they each play a distinct and essential role: trend decomposition captures multi-resolution temporal patterns, conditional diffusion leverages historical context to guide generation, and image-inspired fusion enables effective integration of decomposed signals. The ablation results help isolate their individual contributions to the model’s performance.

The first ablation experiment use consistent configurations: sequence length of 96 time steps for both input and prediction, batch size of 16, learning rate of 0.001 with Adam optimizer, and training for 50 epochs. The models are evaluated using MSE, MAE, and RMSE metrics on an 80/20 train-test split. This experimental design allows us to quantify the individual contribution of each component while maintaining computational efficiency for rapid iteration.

\subsubsection{First Ablation Study Result}The results of the ablation experiments, summarized in \autoref{tab:ablation}, reveal the importance of each component in MR-CDM. To verify the effectiveness of our design, we conducted a systematic ablation study on the LUFL features of the ETTh1 dataset. The trend decomposition module proves essential: removing it (Baseline-NoDecomposition) leads to an 89.6\% performance degradation compared to the FullModel, confirming that multi-scale decomposition is crucial for capturing complex temporal patterns. Even more dramatically, the UnconditionalDiffusion variant—where historical trends are not used to condition the diffusion process—suffers a 1280\% drop in performance, strongly highlighting the necessity of conditioning on coarse-grained historical information for accurate and stable forecasting.

The NoImageFusion variant, which replaces our proposed image-inspired fusion with simple feature concatenation, performs better than the baseline but still falls significantly short of the full model. This demonstrates that naive concatenation fails to exploit the rich cross-scale correlations among decomposed trends, whereas our smart fusion mechanism effectively integrates multi-resolution features. Importantly, the removal of any single component causes a substantial performance decline that cannot be compensated by the remaining modules, validating the complementary nature of our MR-CDM architecture. Together, these components form a synergistic system, enabling the complete model to achieve the best results across all metrics (MSE: 1.4842, MAE: 0.9650, RMSE: 1.2183), thereby fully verifying the effectiveness of our approach.

Experiments demonstrate that MR-CDM significantly outperforms baselines. Ablation studies attribute this gain to three core mechanisms: multi-scale trend decomposition disentangles short-, mid-, and long-term components for independent feature learning, surpassing ARIMA's linearity,LSTM's monolithic encoding and CSDI's instability; conditional diffusion injects historical context to ensure temporal continuity, yielding higher accuracy than unconditional generation; and a multi-path hierarchical reconstructor leverages parallel pathways (hierarchical recovery, cross-scale attention, adaptive weighting) to precisely restore details, avoiding LSTM's gradient vanishing and information loss. Crucially, the complete framework exhibits strong synergistic effects, where the integrated performance exceeds the sum of individual contributions, validating the necessity of the proposed architecture.
\begin{table}[h]
\centering
\caption{Ablation Studies on ETTh1.}
\label{tab:ablation}
\begin{tabular}{lccc}
\hline
Model-Variant & MSE & MAE & RMSE \\ \hline
Baseline-NoDecomposition & {14.2177} & {3.6753} & {3.7706} \\
UnconditionalDiffusion & {20.4822} & {4.4252} & {4.5257} \\
NoImageFusion & {13.0950} & {3.5406} & {3.6187}\\
FullModel & {1.4842} & {0.9650} & {1.2183}\\
\hline
\end{tabular}
\end{table}

\begin{figure}[htbp]
	\raggedleft           
	\includegraphics[width=\linewidth]{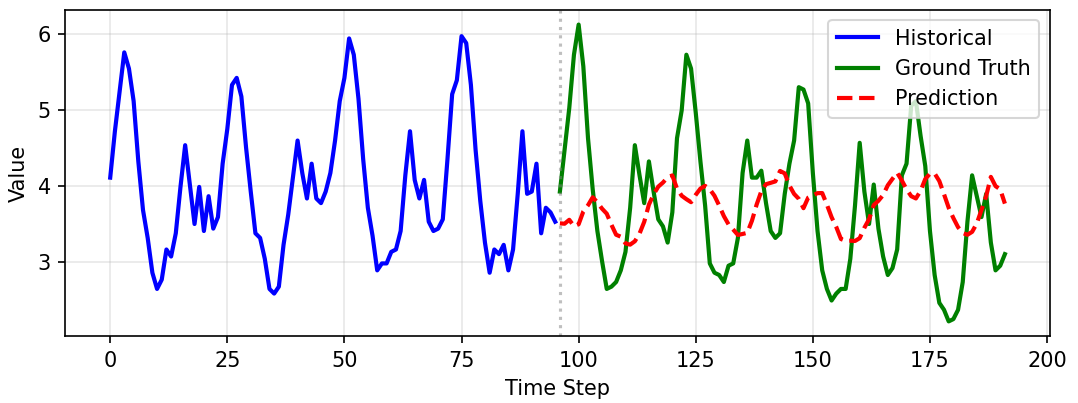}
	\caption{Prediction Result Based on MR- CDM model without conditional information.}
	\label{fig:unconditional}
\end{figure}

To systematically evaluate the role of the conditional information in the proposed diffusion model, we designed a set of ablation experiments, in which all external condition inputs (such as time features, covariates, or historical context guidance) were removed, and the diffusion process itself was solely relied upon for unconditional prediction of the time series. \autoref{fig:unconditional} shows the prediction results of the model on the ETTh1 dataset under this ablation setting. 

Compared with the MR-CDM with conditional information introduced in the main experiment, the unconditional diffusion model can roughly capture the overall trend of the sequence, but it is significantly deficient in detail modeling, phase alignment, and long-term dynamic evolution, manifested as overly smooth prediction curves, delayed peak responses, and distortion of local fluctuations.These results fully demonstrate that the introduced conditional mechanism plays a crucial role in enhancing the expression ability and temporal consistency of the diffusion model in complex time series prediction tasks.

To visually demonstrate the crucial role of conditional information in the diffusion model, \autoref{fig:result} and \autoref{fig:unconditional}  respectively show the performance of the unconditional diffusion model and the proposed conditional diffusion model in the same time series prediction task. Here, the blue solid line represents the historical observed sequence (0–95 steps) as input, the green solid line represents the true values (Ground Truth, 96–191 steps), and the red dashed line represents the model's prediction results. From the figures, it can be seen that under the unconditional setting, although the model can capture the overall trend, there are significant deviations in terms of detail fluctuations, peak responses, and temporal alignment, especially in the rapidly changing areas, indicating its lack of the ability to precisely model the temporal dynamic structure. 

In contrast, after introducing conditional information, the model can more accurately track the intense fluctuations and local features of the true signal, and the prediction curve closely matches the true values, significantly enhancing the ability to restore short-term fluctuations and maintain long-term consistency. This comparison fully validates the effectiveness of the conditional mechanism in guiding the diffusion process and enhancing time-dependent modeling, further highlighting the superior performance of the proposed method in complex time series prediction tasks.

\begin{table}[h]
\centering
\caption{Ablation Studies on Synthetic ETTh1.}
\label{tab:ablation_synthetic}
\begin{tabular}{lccc}
\hline
Model-Variant & MSE & MAE & RMSE \\ \hline
Baseline-NoDecomposition & {108.5863} & {8.0619} & {10.4204} \\
UnconditionalDiffusion & {116.4437} & {8.3155} & {10.7909} \\
NoImageFusion & {111.8180} & {8.2349} & {10.5744}\\
FullModel & {1.2544} & {0.9080} & {1.1200 }\\
\hline
\end{tabular}
\end{table}
As shown in \autoref{tab:ablation_synthetic},we further conducted ablation studies on the synthetic dataset, and the results consistently demonstrate that each component of MR-CDM plays a vital role: multi-scale trend decomposition effectively captures hierarchical temporal patterns, conditional diffusion leverages historical context for accurate generation, and the image-based fusion mechanism enables meaningful integration of multi-scale features. Removing any one of these components leads to a significant performance drop, confirming that the full architecture is necessary and well-designed.

\subsection{Second Ablation Study}
\label{sec:ablation2}
To systematically evaluate the structural effectiveness of the proposed MR-CDM in multi-scale time-series modeling, we conduct an ablation study on the trend decomposition module. The experiments are performed on the LUFL variable of the ETTh1 dataset under a unified setting of prediction length, with MSE, MAE, and RMSE adopted as evaluation metrics. Specifically, three configurations are compared: the full MR-CDM model, MR-CDM w/o Trend1 (removing the low-frequency trend component), and MR-CDM w/o Trend3 (removing the high-frequency component). The reason for not removing Trend2 is that it is not the "highest frequency" nor the "smoothest", but somewhere in between, mainly carrying medium-term structural information.

The results \autoref{tab:ablation_trendDecomposition} show that the full model achieves the best performance across all three metrics. Removing either trend component leads to a noticeable degradation in prediction accuracy, with a larger drop observed when Trend3 is removed. This indicates that the high-frequency component is critical for short-term fluctuation modeling, while the low-frequency component remains indispensable for preserving global trend dynamics. Overall, the proposed multi-scale trend decomposition mechanism enables collaborative modeling of short-term disturbances and long-term evolution within a unified framework, thereby improving model robustness and generalization in practical forecasting scenarios.

\begin{table}[h]
\centering
\caption{Trend Decomposition Ablation Result.}
\label{tab:ablation_trendDecomposition}
\begin{tabular}{lccc}
\hline
Model-Variant & MSE & MAE & RMSE \\ \hline
MR-CDM w/o Trend1 & {1.9342} & {1.1247} & {1.3907} \\
MR-CDM w/o Trend3 & {2.2769} & {1.2678} & {1.5089} \\
FullModel & {1.4842} & {0.9650} & {1.2183}\\
\hline
\end{tabular}
\end{table}

\subsection{Input Length Sensitivity Analysis}\label{sec:ablation3}
To further investigate the effect of historical input length on the forecasting performance of MR-CDM, we conduct an input length sensitivity analysis on the LUFL variable of the ETTh1 dataset. The prediction length is fixed at 96, while the input sequence length $\text{Seq\_Len}$ is varied across 48, 96, and 192. As shown in the \autoref{tab:Sensitivity Analysis}, all three error metrics decrease monotonically as the input length increases, indicating that longer historical sequences provide richer temporal context for the model.Longer input sequences contain more complete periodic structures and trend dynamics, enabling the multi-scale decomposition module to extract each frequency component more accurately and reducing decomposition errors caused by insufficient historical context.  This enables the multi-scale trend decomposition module to extract more complete periodic and trend features, thereby improving prediction accuracy. These results confirm that MR-CDM is adaptive to varying input lengths, and suggest that increasing the historical window size in practical deployment can further enhance forecasting performance.

\begin{table}[h]
\centering
\caption{Input Length Sensitivity Results.}
\label{tab:Sensitivity Analysis}
\begin{tabular}{lccc}
\hline
$\text{Seq\_Len}$ & MSE & MAE & RMSE \\ \hline
48 & {1.8998} & {1.1387} & {1.3783} \\
96 & {1.4842} & {0.9650} & {1.2183} \\
192 & {1.2319} & {0.8492} & {1.0965}\\
\hline
\end{tabular}
\end{table}

\section{Generalization to Other Domains}
\label{sec:Generalization}
To further validate the generalization capability and robustness of our model, we extended our evaluation to datasets from distinct domains: weather forecasting and traffic flow prediction. These datasets embody complex non-linear spatiotemporal dynamics characteristic of meteorological variations and urban traffic patterns, respectively, thereby serving as rigorous benchmarks for assessing cross-domain adaptability.In our experiments, the OT metric serves as the ground truth for forecasting tasks on both datasets. Experimental results demonstrate that our model not only retains its performance advantages but also achieves superior prediction accuracy and stability in these diverse scenarios, underscoring its significant potential for broad real-world applications.\autoref{tab:weather} and \autoref{tab:traffic} show the prediction ability among these models.

Weather dataset comprises 21 multivariate time series collected from a meteorological station in Jena, Germany, with a sampling rate of 10 minutes. Characterized by complex seasonal patterns (daily and yearly cycles) and high-frequency noise, this dataset tests the model's robustness in handling non-stationary environmental data and capturing inter-variable dependencies.

\begin{table}[h]
\centering
\caption{Model Performance on Weather Domain.}
\label{tab:weather}
\begin{tabular}{lccc}
\hline
Model & MSE & MAE & RMSE \\ \hline
ARIMA & {191.2356} & {11.6258} & {13.8287} \\
LSTM & {124.7143} & {7.5514} & {11.1675}\\
CSDI & {96.3245} & {6.9438} & {9.8145}\\
MR-CDM & {79.4128} & {6.3487} & {8.9113} \\
\hline
\end{tabular}
\end{table}

Traffic dataset records the road occupancy rates collected from 862 sensors in the San Francisco Bay Area. The data is sampled hourly and exhibits strong daily and weekly periodicity, as well as complex spatial dependencies among sensors. It serves as a rigorous benchmark for evaluating a model's ability to capture recurring patterns and handle high-dimensional multivariate series.
\begin{table}[htbp]
\centering
\caption{Model Performance on Traffic Domain.}
\label{tab:traffic}
\begin{tabular}{lccc}
\hline
Model & MSE & MAE & RMSE \\ \hline
ARIMA & {0.0011} & {0.0237} & {0.0331} \\
LSTM & {0.0007} & {0.0152} & {0.0264}\\
CSDI & {0.0004} & {0.0143} & {0.0200}\\
MR-CDM & {0.0003} & {0.0139} & {0.0173} \\
\hline
\end{tabular}
\end{table}

\section{Multiple-Run Validation}
\label{sec:multiple}
To ensure the robustness and reproducibility of our results, each experiment is repeated three times using distinct random seeds (specifically, 42, 43, and 44). This practice helps account for the inherent stochasticity in model initialization and training dynamics. The corresponding statistical summaries—including mean and standard deviation of key performance metrics across the three runs—are reported in \autoref{tab:multiple-run}(ETTh1) and \autoref{tab:multiple-run-sythetic}(Synthetic ETTh1). These aggregated results provide a more reliable basis for comparison and mitigate the influence of random variation on our conclusions.

\begin{table}[h]
\centering
\caption{Multiple-Run Performance on ETTh1.}
\label{tab:multiple-run}
\begin{tabular}{lccc}
\hline
Model & MSE & MAE & RMSE \\
\hline
ARIMA & {30.0946} & {8.2296} & {5.4858}\\
LSTM &  {13.2883} & {6.2361} & {3.6453}\\
CSDI & {2.7493} & {1.2478} & {1.6581}\\
\hline
\end{tabular}
\end{table}
To ensure the reliability and reproducibility of our experimental results, we conducted multiple-run validation for all baseline models. Specifically, we trained and evaluated ARIMA,LSTM and CSDI models three times with different random seeds, and reported the averaged performance metrics across all runs. This rigorous validation protocol helps mitigate the impact of random initialization and stochastic training processes, providing more robust and statistically reliable comparisons. The results presented in \autoref{tab:multiple-run} show the mean performance across three independent runs. These averaged results demonstrate consistent performance patterns across multiple runs, confirming the stability of baseline model performance and ensuring fair comparison with our proposed method. The relatively small variance across runs (not shown for brevity) further validates the reliability of our experimental setup and the robustness of the reported performance improvements.

\begin{table}[h]
\centering
\caption{Multiple-Run Performance on Synsetic ETTh1.}
\label{tab:multiple-run-sythetic}
\begin{tabular}{lccc}
\hline
Model & MSE & MAE & RMSE \\
\hline
ARIMA & {47.2826} & {7.4434} & {6.8762}\\
LSTM &  {13.6146} & {3.6670} & {3.6897}\\
CSDI & {2.6159} & {1.9632} & {1.6173}\\
\hline
\end{tabular}
\end{table}
Similarly, we also repeated the aforementioned baseline experiments on the synthetic dataset three times, using different random seeds to account for stochastic variations in model initialization and training. The results across all runs demonstrate consistent performance for ARIMA, LSTM and CSDI baseline models, confirming their stability on this controlled dataset. As shown in the corresponding evaluation metrics reported in the relevant \autoref{tab:multiple-run-sythetic}, the low variance across repetitions further validates the reliability of these baselines under synthetic conditions. These findings not only reinforce the robustness of our experimental setup but also provide a solid foundation for comparing more advanced methods, thereby supporting our overall analysis and conclusions.

Experiments demonstrate that MR-CDM significantly outperforms baselines. Experiments studies attribute this gain to three core mechanisms: multi-scale trend decomposition disentangles short-, mid-, and long-term components for independent feature learning, surpassing ARIMA’s linearity,LSTM’s monolithic encoding and CSDI's instability; conditional diffusion injects historical context to ensure temporal continuity, yielding higher accuracy than unconditional generation; and a multi-path hierarchical reconstructor leverages parallel pathways (hierarchical recovery, cross-scale attention, adaptive weighting) to precisely restore details, avoiding LSTM’s gradient vanishing and information loss. Crucially, the complete framework exhibits strong synergistic effects, where the integrated performance exceeds the sum of individual contributions, validating the necessity of the proposed architecture.

\section{Background}\label{sec:background}

\textbf{Diffusion Models.}
Diffusion models~\cite{NEURIPS2020_4c5bcfec} consist of a forward noising and a backward denoising process. 
\textit{Forward Diffusion} gradually adds Gaussian noise over $K$ steps. The closed-form expression for step $k$ is:
\begin{equation}
    \mathbf{x}^k = \sqrt{\bar{\alpha}_k}\mathbf{x}^0 + \sqrt{1-\bar{\alpha}_k}\epsilon, \quad \epsilon \sim \mathcal{N}(\mathbf{0},\mathbf{I}),
\end{equation}
where $\bar{\alpha}_k = \prod_{s=1}^k (1-\beta_s)$.
\textit{Reverse Denoising} recovers clean data from noise, typically by predicting the noise $\epsilon$ or the clean data $\mathbf{x}^0$ via:
\begin{equation}
    \mathcal{L}_\epsilon = \mathbb{E} \|\epsilon - \epsilon_\theta(\mathbf{x}^k, k)\|^2 \quad \text{or} \quad \mathcal{L}_{\mathbf{x}} = \mathbb{E} \|\mathbf{x}^0 - \mathbf{x}_\theta(\mathbf{x}^k, k)\|^2.
\end{equation}

\textbf{Conditional Diffusion Models.}
For time series prediction, the denoising process is conditioned on historical context $\mathbf{c} = \mathcal{F}(\mathbf{x}_{-L+1:0}^0)$. The conditional distribution is defined as:
\begin{equation}
    p_\theta(\mathbf{x}_{1:H}^{0:K} \mid \mathbf{c}) = p_\theta(\mathbf{x}_{1:H}^K) \prod_{k=1}^K p_\theta(\mathbf{x}_{1:H}^{k-1} \mid \mathbf{x}_{1:H}^k, \mathbf{c}),
\end{equation}
where the mean $\mu_\theta(\mathbf{x}^k, k \mid \mathbf{c})$ is computed by leveraging both the noisy input and the conditioning context $\mathbf{c}$.

\textbf{Hierarchical Trend Decomposition (HTD).}
HTD~\cite{shen_multi-resolution_2024} decomposes time series into multi-scale trends. Given a series $X_0$, trend components are extracted sequentially:
\begin{equation}\label{eq:treand_extraction}
    X_s = \text{AvgPool}(\text{Padding}(X_{s-1}), \tau_s), \;\;\; s=1,\dots,S-1,
\end{equation}
where $\text{AvgPool}$ is average pooling and $\tau_s$ (increasing with $s$) controls the smoothing kernel size, enabling extraction of progressively coarser trends.

\textbf{Time Series to Image Transforms.}
Time series are mapped to images using invertible transforms to exploit spatial inductive biases~\cite{takens2006detecting,DBLP:conf/icassp/GriffinL83}:
\begin{itemize}
    \item \textbf{Delay Embedding:} For a univariate series $x_{1:L}$, it constructs a matrix $X \in \mathbb{R}^{n \times q}$ where $n$ is the embedding dimension and $q$ is the number of time steps. Adaptive variants dynamically adjust $m$ and $n$ to capture complex dynamics.
    \item \textbf{Short Time Fourier Transform (STFT):} Maps the signal to the frequency domain using a sliding window. For input $\mathbf{x} \in \mathbb{R}^{L \times K}$, STFT outputs $\mathbf{x}_{\text{img}} \in \mathbb{R}^{2K \times H \times W}$ (storing real/imaginary parts). It is invertible with negligible information loss.
\end{itemize}

\section{Additional datasets / synthetic results}
\label{sec:additional}
To further validate robustness and generalization, we conduct the same experiments on a synthetic dataset with similar statistical properties, multi-scale periodicity, and correlations as ETTh1, as shown in \autoref{tab:main_results_systhetic}. The consistent improvements on both real and synthetic data demonstrate that MR-CDM does not overfit specific datasets but effectively captures underlying temporal dynamics. This validates the generalizability of combining multi-scale decomposition with conditional diffusion for time series forecasting.

\begin{table}[h]
\centering
\caption{Performance comparison on Sythetic Dataset.}
\label{tab:main_results_systhetic}
\begin{tabular}{lccc}
\hline
Model & MSE & MAE & RMSE\\
\hline
ARIMA & {39.8131} & {8.3046} & {6.3097}\\
LSTM & {12.7953} & {8.2361} & {3.5770}\\
CSDI & {1.6749} & {0.8741} & {1.2941}\\
MR-CDM & {1.2544} & {0.9080} & {1.1200}\\
\hline
\end{tabular}
\end{table}

\section{Additional qualitative comparison}
\subsection{CSDI}
\label{sec:aqc}
\begin{figure}
	\centering           
	\includegraphics[width=0.8\linewidth]{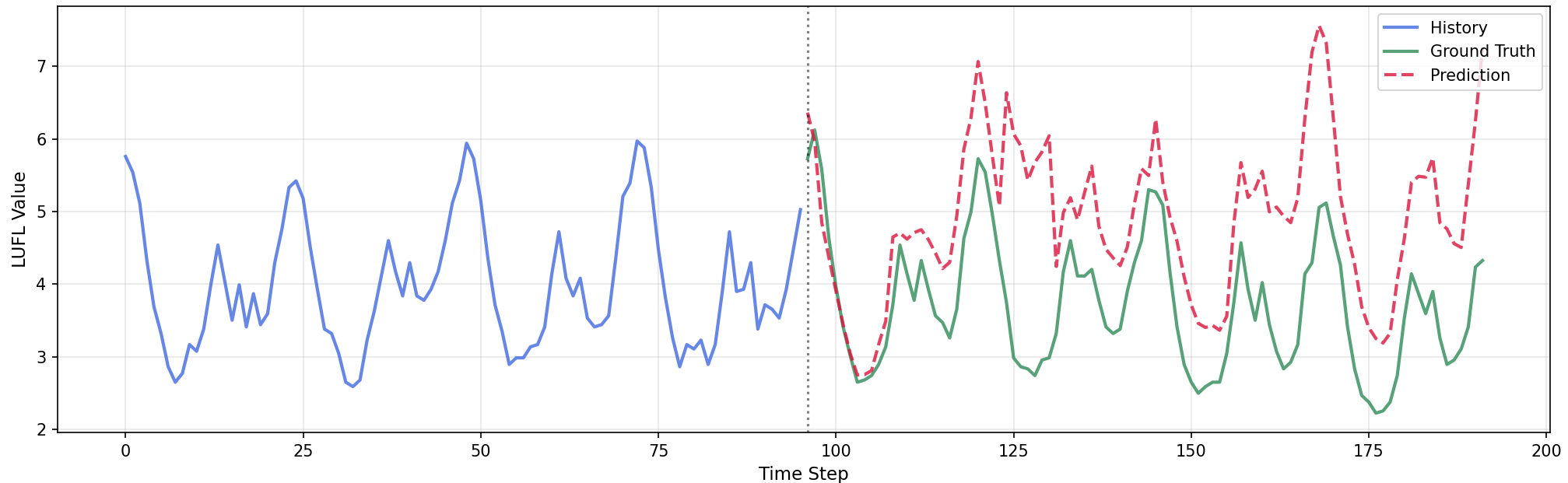}
	\caption{The Prediction Result Based on CSDI Model.}
	\label{fig:csdi_result}
\end{figure}

Experimental results on ETTh1 further confirm that MR-CDM outperforms CSDI, as shown in \autoref{fig:csdi_result}. This advantage arises from key architectural differences. MR-CDM explicitly decomposes time series into multi-scale components and applies tailored strategies such as delay embedding and STFT, whereas CSDI relies on a unified convolution that limits multi-scale representation. By transforming time series into the image domain, MR-CDM better leverages diffusion-based generation. In addition, its multi-path hierarchical reconstructor captures inter-scale dependencies more effectively than CSDI’s single-path decoder, leading to improved feature learning and prediction accuracy.

\subsection{MR-CDM}
\label{sec:aqc1}
\begin{figure}
	\centering           
	\includegraphics[width=0.7\linewidth]{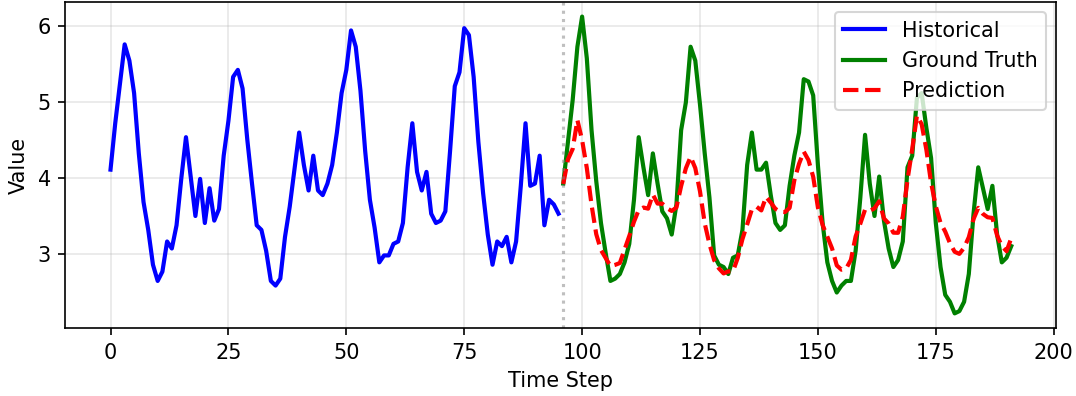}
	\caption{The Prediction Result Based on MR-CDM Model.}
	\label{fig:result}
\end{figure}

As illustrated in \autoref{fig:result}, we evaluate the prediction performance of MR-CDM on long-term LUFL forecasting using the ETTh1 dataset. The input sequence consists of the first 96 time steps (0–95), while the subsequent 96 steps (96–191) form the prediction horizon, covering a total duration of 192 hours (8 days) with hourly resolution.

\section{Experimental Environment}
\label{sec:Experimental Env}
All experiments were conducted on a high-performance workstation configured with an Intel Xeon Silver 4110 CPU, 256 GB of DDR4 RAM, and two NVIDIA TITAN RTX GPUs, each equipped with 24 GB of dedicated memory. The software stack was built on Ubuntu 20.04 LTS and included Python 3.11, PyTorch 2.0.1, and CUDA 11.8 to enable efficient GPU-accelerated training and inference. This setup ensured consistent and reproducible experimental conditions across all evaluations.

\section{Dataset Synthesis}\label{sec:DS}
To demonstrate the generalization capability in similar dataset of our model, we generated a synthetic dataset using a multi-component time series synthesis approach. The synthetic data preserves the statistical properties (mean, standard deviation, and value ranges) of the original ETTh1 dataset while incorporating realistic temporal patterns including daily cycles (24-hour periodicity), weekly cycles (7-day periodicity), long-term trends, and Gaussian noise. Inter-variable correlations were maintained through correlated signal generation, and daytime load enhancement was applied to simulate realistic power consumption patterns. 

\section{Details of Feature Processing}\label{sec:FP}
In the decomposition stage, we employ three moving average filters with window sizes of $[5, 25, 51]$, corresponding to temporal scales of approximately 5 hours, 1 day, and 2 days, respectively. This configuration enables the capture of multi-level features ranging from short-term fluctuations to long-term trends. Specifically, the filter with window size 5 (MA-5) extracts high-frequency short-term fluctuations, MA-25 identifies daily periodic patterns, and MA-51 captures long-term trends. This hierarchical design ensures that features at distinct temporal scales are disentangled and can be subjected to differentiated processing strategies.
\textbf{Adaptive Image Transformation.} 
For the components $\text{Trend}_1$, $\text{Trend}_2$, and the $\text{Residual}$, we utilize the delay embedding method to transform 1D time series into $32 \times 32$ 2D images, setting the delay parameter $\tau=3$ and the embedding dimension $d=32$. Here, $\tau=3$ effectively captures temporal dependencies over a 3-hour span, while $d=32$ aligns with the base resolution of the U-Net architecture, thereby avoiding unnecessary upsampling or downsampling operations. Conversely, for $\text{Trend}_3$, we apply the Short-Time Fourier Transform (STFT) to convert the signal into the time-frequency domain, configured with $\text{n\_fft}=64$ and $\text{hop\_length}=16$. The choice of $\text{n\_fft}=64$ yields 32 frequency components sufficient for identifying periodic characteristics, while $\text{hop\_length}=16$ ensures a 75\% window overlap, guaranteeing temporal continuity and information integrity.

\textbf{Image Fusion.} 
In the fusion stage, we adopt a channel concatenation strategy to merge the images of the four components into a single fused tensor with 35 channels (structured as $7+7+14+7$). This design preserves the complete information of all components, facilitating subsequent hierarchical reconstruction.
\noindent